\documentclass[sigconf]{acmart}

\usepackage{booktabs} 
\usepackage{makecell} 
\usepackage{balance}
\usepackage{makecell}
\usepackage{algorithm}
\usepackage{algpseudocode}

\copyrightyear{2020}
\acmYear{2020}
\acmConference[ISMSI 2020]{ISMSI 2020}{March, 2020}{ISMSI}
\acmBooktitle{ISMSI 2020, Thimphu, Kingdom of Bhutan}

\begin{document}
\title[Population-based metaheuristics for Association Rule Text Mining]{Population-based metaheuristics for Association Rule Text Mining}

\author{Iztok Fister Jr.}
\affiliation{
  \institution{University of Maribor}
  \streetaddress{Maribor, Slovenia}
  \city{Maribor} 
  \country{Slovenia} 
}
\email{iztok.fister1@um.si}

\author{Suash Deb}
\affiliation{
  \institution{Victoria University, Decision Sciences and Modeling Program}
  \city{Melbourne, Australia}
}
\affiliation{
  \institution{IT \& educational Consultant, Ranchi, Jharkhand, India}
}

\email{suashdeb@gmail.com}  

\author{Iztok Fister}
\affiliation{
  \institution{University of Maribor}
  \city{Maribor} 
  \country{Slovenia} 
}
\email{iztok.fister@um.si}

\renewcommand{\shortauthors}{I. Fister Jr. et al.}

\begin{abstract}
Nowadays, the majority of data on the Internet is held in an unstructured format, like websites and e-mails. The importance of analyzing these data has been growing day by day. Similar to data mining on structured data, text mining methods for handling unstructured data have also received increasing attention from the research community. The paper deals with the  problem of Association Rule Text Mining. To solve the problem, the PSO-ARTM method was proposed, that consists of three steps: Text preprocessing, Association Rule Text Mining using population-based metaheuristics, and text postprocessing. The method  was applied to a transaction database obtained from professional triathlon athletes' blogs and news posted on their websites. The obtained results reveal that the proposed method is suitable for Association Rule Text Mining and, therefore, offers a promising way for further development.
\end{abstract}


\maketitle

\section{Introduction}
Stochastic population-based nature-inspired metaheuristics offer a very effective way for Association Rule Mining (ARM). They are stochastic in their nature, and do not discover association rules using an  exhaustive search as the other classical methods do. Numerous population-based methods exist for ARM that were developed in the past years. According to the literature review, most of the existing methods are intended for mining categorical features that are stored in transaction databases. On the other hand, some methods also exist that can deal with numerical data. Actually, this kind of mining is also called Numerical Association Rule Mining (NARM)~\cite{altay2019performance}. 

Some examples of the ARM based on stochastic population-based nature-inspired algorithms include methods like MODENAR~\cite{alatas2008modenar}, ARMGA~\cite{qodmanan2011multi}, and ARM-DE~\cite{Fister2018ARM}. MODENAR is an example of a very efficient multi-objective Differential Evolution for mining numeric association rules, while ARMGA is a Genetic Algorithm for discovering association rules, where there is no necessary to specify the minimum support and minimum confidence by users. On the other hand, ARM-DE is a new approach for NARM problems, based on Differential Evolution.

In contrast, there is a lack of works for discovering the association rules in text~\cite{manimaran2013survey}.  Association Rule Text Mining (ARTM) results in many practical applications, e.g., for building  text classifiers~\cite{antonie2002text,rahman2010text}. On the other hand, methods of generating knowledge discovery in text mining using association rule extraction are helpful for the users who are able to find accurate and important knowledge quicker than browsing through the text manually~\cite{kulkarni2016knowledge}.  ARTM is also very interesting in the Medical domain~\cite{boytcheva2018indirect}. In the paper~\cite{kamruzzaman2010text}, association rules are used to derive a feature set from pre-classified text documents. Interestingly, the authors of a survey paper~\cite{manimaran2013survey} discovered that the Apriori algorithm is also suitable for ARTM, and utilized mostly in various domains, especially in the domain of Medical Care~\cite{manimaran2013survey}. 

To the authors` knowledge, no  methods exist for ARTM that are based fully on stochastic population-based nature-inspired metaheuristics. In this paper, we tackle the problem of ARTM using  Particle Swarm Optimization (PSO)~\cite{kennedy1995particle}. We would like to get some answers on the following questions:
\begin{itemize}
\item Are stochastic population-based nature-inspired metaheuristic algorithms suitable for ARTM?
\item Can we find a viable interpretation of discovered association rules in text?
\item Is there a bright way of developing these algorithms in the future?
\end{itemize}

The paper is structured as follows: Sec.~\ref{arm-features} depicts some features of ARTM, as well as outlines the main differences between conventional ARM tasks, while Sec.~\ref{proposed} deals with a detailed description of the proposed method. Experiments and results are presented in Sec.~\ref{experiments}, while the paper is concluded with Sec.~\ref{conclusion}, where directions for the future work are outlined. 
 
\section{Basic information}
\label{arm-features}
This section is focused on the  background information necessary for understanding the subject that follows. At first, the problem of discovering association rules is illustrated in detail, followed by the  basics of the PSO algorithm.

\subsection{Association Rule Mining}
ARM can formally be defined as follows: Let us assume a set of objects $O=\{o_1,\ldots,o_M\}$ and transaction dataset $T_D=\{T\}$ are given, where each transaction $T$ is a subset of objects $T\subseteq O$. Then, an association rule is defined as an implication:
\begin{equation}
    X\Rightarrow Y,
\end{equation}
where $X\subset O$, $Y\subset O$, and $X\cap Y=\emptyset$. In order to estimate the quality of mined association rule, two measures are defined: A confidence and a support. The confidence is defined as:
\begin{equation}
    conf(X\Rightarrow Y)=\frac{n(X\cup Y)}{n(X)},
\end{equation}
while the support as:
\begin{equation}
    supp(X\Rightarrow Y)=\frac{n(X\cup Y)}{N},
\end{equation}
where function $n(.)$ calculates the number of repetitions of a particular rule within $D_T$, and $N$ is the total number of transactions in $D_T$. Let us emphasize that two additional variables are defined, i.e., the minimum confidence $C_{min}$ and the minimum support $S_{min}$. These variables denote a threshold value limiting the particular association rule with lower confidence and support from being taken into consideration.

\subsection{Basics of the PSO algorithm}
Particle Swarm Optimization (PSO) is a member of an SI-based algorithm family that was developed by Eberhard and Kennedy in 1995~\cite{kennedy1995particle}. It is inspired by the social behavior of bird flocking and fish schooling. This algorithm works with a swarm (i.e., a population) of particles representing candidate solutions $\mathbf{x}^{(t)}_i$ of the problem to be solved. The particles fly virtually through the problem space, and are attracted by regions reached with food. When the particles are located in the vicinity of these regions, they are rewarded with the better values of fitness function by the algorithm. 

Interestingly, the PSO algorithm exploits usage of an  additional memory, where the particle's personal best $\mathbf{p}^{(t)}_i$ as well as the swarm's global best $\mathbf{g}^{(t)}$ locations in the search space are saved. In each time step $t$ (i.e., generation), all particles change their velocities $\mathbf{v}^{(t)}_i$ towards their  personal and global best locations according to the following mathematical formula:
\footnotesize
\begin{equation}
\begin{aligned}
\mathbf{v}^{(t+1)}_i &= \mathbf{v}^{(t)}_i+C_1\cdot \text{rand}(0,1)\cdot\left(\mathbf{g}^{(t)}-\mathbf{x}^{(t)}_i\right)+C_2\cdot \text{rand}(0,1)\cdot\left(\mathbf{p}^{(t)}_i-\mathbf{x}^{(t)}_i\right), \\ 
\mathbf{x}^{(t+1)}_i &= \mathbf{x}^{(t)}_i+\mathbf{v}^{(t)}_i,
\end{aligned}
\label{eq:pso}
\end{equation}
\normalsize
where $C_1$ and $C_2$ present social and cognitive weights typically initialized to 2, and $\text{rand}(0,1)$ is a random value drawn from uniform distribution in the interval $[0,1]$.

The pseudo-code of the original PSO is illustrated in Algorithm~\ref{alg:pso},
\begin{algorithm}[htb]
\caption{The original PSO algorithm}
\label{alg:pso}
\begin{algorithmic}[1]
\Procedure{ParticleSwarmOptimization}{}
\State $t\gets 0$;
\State $P^{(t)}\gets$\textsc{Initialize}; \Comment {initialization of population}
\While {$\mathbf{not}$ \textsc{TerminationConditionMeet}}
\ForAll {$\textbf{x}^{(t)}_i\in P^{(t)}$}
\State $f^{(t)}_{i}$ = \textsc{Evaluate}($\mathbf{x}^{(t)}_i$); \Comment {evaluation of candidate}
\If {$f^{(t)}_{i} \leq f^{(t)}_{\mathit{best}_i}$}
\State $\mathbf{p}^{(t)}_{i}=\mathbf{x}^{(t)}_{i}$; $f^{(t)}_{\mathit{best}_i}=f^{(t)}_{i}$;
\EndIf \Comment {preserve the local best solution} 
\If {$f^{(t)}_{i} \leq f^{(t)}_{\mathit{best}}$}
\State $\mathbf{g}^{(t)}=\mathbf{x}^{(t)}_{i}$; $f^{(t)}_{\mathit{best}}=f^{(t)}_{i}$;
\EndIf \Comment {preserve the global best solution }
\State $\mathbf{x}^{(t)}_i$ = \textsc{Move}($\mathbf{x}^{(t)}_i$); \Comment{move candidate w.r.t. Eq.~(\ref{eq:pso})}
\EndFor
\State $t=t+1$;
\EndWhile 
\EndProcedure
\end{algorithmic}
\end{algorithm}
from which it can be seen that the PSO is distinguished from the classical EAs by three specialties: 
\begin{itemize}
\item does not have survivor selection,
\item does not have the crossover operator,
\item the mutation operator is replaced by the move operator changing each element of particle $\mathbf{x}^{(t)}_i$ with the probability of mutation $p_m=1.0$,
\item does not have the selection operator.
\end{itemize}
Let us mention that the selection is implemented in the PSO implicitly, i.e., by improving the personal best solution permanently. However, when this improving is not possible anymore, the algorithm gets stuck in the local optima.

\section{Proposed method}
\label{proposed}
At a glance, there is no simple analogy between text and market basket analysis that is a good example of an ARM task. Items in a market basket are self-contained, and well organized in transaction databases, where they are easy to employ by algorithms for ARM. However, working with pure text is a totally different story, because here, information is hidden in unstructured text. As stated by Brownlee~\cite{brownlee2014machine}, "a problem with modeling text is that it is messy, and techniques like machine learning algorithms prefer well defined fixed-length inputs and outputs". 

Consequently, the unstructured text needs complex preprocessing treatment, where those unstructured data must be transformed into a structured transaction database. Obviously, the transaction database is appropriate for discovering the association rules using stochastic population-based nature-inspired algorithms. For the needs of this study, the PSO algorithm was employed for solving the ARTM (i.e., PSO-ARTM).  

In general, the proposed PSO-ARTM consists of the following three phases:
\begin{itemize}
\item text preprocessing,
\item optimization,
\item postprocessing.
\end{itemize}
In the remainder of the paper, the aforementioned phases are presented in detail.

\subsection{Text preprocessing}
The purpose of this phase is to generate a transaction database from the raw data obtained from triathlon athletes' blogs and news posted on their websites, and consists of three steps, as follows:
\begin{itemize}
    \item tokenizing,
    \item stop word removal,
    \item term frequencies` calculation.
\end{itemize}
Punctuation marks are removed in the first step. As a result, only words delimited by space remain in the document. Some words, like definite and indefinite articles (e.g., the, a, an), connective words (e.g., and, also, then), conjunctions (e.g., but, when, because), and verbs (e.g., is, done), represent the so-called stop words, and must be removed in the second step. The result of this step is a sequence of terms. The terms undergo term frequency calculation, where occurrences are not only determined, but also weighted. Thus, a Term Frequency/Inverse Term Frequency (TF/ITF) weighting scheme is used that penalizes the rare occurring terms with higher weights.  

The TF/ITF weighting scheme is defined as follows: For a given term $z_j$, for $j=1,\ldots ,M$, occurring in document $d_i$, for $i=1,\ldots, N$, the term frequency is expressed as:
\begin{equation}
    \mathit{TF}_{i,j}=\frac{n(d_i,w_j)}{|d_i|},
\end{equation}
where $n(d_i,w_j)$ denotes the number of occurrences of term $w_j$ in document $d_i$, and $|d_i|$ is the total number of terms in document $D_i$. On the other hand, the inverse term frequency is expressed as:
\begin{equation}
    \mathit{ITF}_j=\left|\log \frac{n(d|w_j)}{N}\right|,
\end{equation}
where $n(d|w_j)$ denotes the number of document $d$ containing term $w_j$, and $N$ is the total number of documents.

Furthermore, the weighted frequency of term $z_j$ in document $d_i$ is represented as a vector of weights $\mathbf{w}_i=\{w_{i,1},\ldots,w_{i,n}\}$, where each element $w_{i,j}$ is expressed as:
\begin{equation}
    w_{i,j}=\mathit{TF}_{i,j}\cdot \mathit{ITF}_j,\quad\text{for}~j=1,\ldots,n.
\end{equation}
Finally, the transaction database is generated from the relevant documents by moving each vector $\mathbf{w}_i$, representing weighted frequencies for all terms in the corresponding document, to a transaction in $D_T$. In this way, the transaction database is very similar to the market basket, except that the weighted frequencies are put into $D_T$ instead of the value of one.

\subsection{Optimization}
An ARTM problem can be defined formally as follows: Let us assume a set of documents $D=\{d_1,\ldots,d_N\}$ and set of terms $Z=\{z_1,\ldots,z_M\}$ are given, where $N$ denotes the maximum number of documents, and $M$ the maximum number of terms, respectively, to which also the matrix of weights $\mathbf{W}$ of dimension $N\times M$ is assigned, where each element $w_{i,j}$ represents a frequency weight of term $z_j$ in document $d_i$, calculated according the TF-ITF weighting scheme. Then, the task of optimization is to select the binary vector $\mathbf{y}=(y_1,\ldots,y_M)^T$, determining the presence or absence of the corresponding term in the solution, such that 
the scalar product 
\begin{equation}
    \mathit{AWS}=\sum_{j=1}^{M}\sum_{i=1}^{N}{w_{i,j}\cdot y_j}
    \label{eq:aws}
\end{equation}
subject to
\begin{equation}
    \sum_{j=1}^M y_j\leq K,
    \label{eq:ineq}
\end{equation}
is maximum. Let us mention that variable $K$ denotes the maximum number of terms in association rule. Actually, the selected elements of vector $\mathbf{y}$ form the set $Y=\{y_j|y_j=1,~\text{for}~j=1,\ldots,M\}$ that is a subset of $Z$, in other words $Y\subset Z$. Let us notice that the values of vector are initially set to zero.  

The problem is solved using the PSO algorithm, that needs the following modifications referred to three components of the algorithm: (1) Representation of individuals, (2) Genotype-phenotype mapping, and (3) Evaluation function. In the remainder of the paper, the aforementioned modifications are discussed in detail.

\subsubsection{Representation of individuals}
The candidate solutions in the PSO algorithm are represented as real-valued vectors
\begin{equation}
    x_i^{(t)}=(x_{i,1}^{(t)},\ldots,x_{i,K}^{(t)},x_{i,K+1}^{(t)}),\quad \text{for}~i=1,\ldots,\mathit{Np},
\end{equation}
where $x_{i,j}^{(t)}\in [0,1]$ for $j=1,\ldots,K$ encodes the selected terms in association rule, $K$ the maximum number of terms in the particular association rule, $\mathit{Np}$ is the population size, and $t$ the generation counter. However, the last element of vector $x_{i,K+1}^{(t)}$ determines the cut point between antecedent and consequence in the rule. 

\subsubsection{Genotype-phenotype mapping}
Each solution encodes a definite association rule in the genotype space that needs to be mapped into phenotype space before evaluation. This mapping is performed according to the following equation:
\begin{equation}
    y_j^{(t)}=\left \{ \begin{matrix}
    1, & \text{if}~\left\lfloor \frac{x_{i,j}^{(t)}}{K} \right\rfloor=j,\\
    0, & \text{otherwise},
    \end{matrix} \right.\quad \text{for}~j=1,\ldots,K.
    \label{eq:mapping}
\end{equation}
The cut point denoted by the last element $x_{i,K+1}^{(t)}$ is calculated according to the following equation:
\begin{equation}
    \mathit{cp}=\left\lfloor \frac{x_{i,j}^{(t)}}{K-1} \right\rfloor,
\end{equation}
which selects one of the $K-1$ cut points between $K$ elements of the association rule.

Let us mention that the mapping in Eq.~(\ref{eq:mapping}) does not ensure the injective mapping of each element of a vector to the specific term. This means that it is possible that more elements of vector $x_{i,j}^{(t)}$ are mapped to the same term $y_j^{(t)}$. In this case, the number of terms in the association rule is less than $K$, but this is admissible according to Eq.~(\ref{eq:ineq}).

As a result of genotype-phenotype mapping, the association rule $X\Rightarrow Y$ is obtained, where the cut point delineates the antecedent part from the consequence.

\subsubsection{Evaluation function}
Evaluation function in the PSO algorithm for ARTM estimates the quality of the association rule according to the following equation:
\small
\begin{equation}
    f(X \Rightarrow Y)=\frac{\alpha\cdot\mathit{supp}(X\Rightarrow Y)+\beta\cdot\mathit{conf}(X\Rightarrow Y)+\gamma\cdot\mathit{AWS}}{\alpha+\beta+\gamma},     
    \label{eq:fit}
\end{equation}
\normalsize
where $\alpha$, $\beta$, and $\gamma$ represent the weights of the particular terms. In our study, the values of weights are fixed to one. This means that each term in Eq.~(\ref{eq:fit}) is treated equally. Let us mention that the value of the fitness function needs to be maximized, in other words $f^{*}(X\Rightarrow Y)=\max f(X\Rightarrow Y)$.

\subsection{Postprocessing}
The aim of this step is the interpretation of the results. Typically, the results of the Association Rule Mining are illustrated in  tabular form. This way of presentation is also applied in our study.

\section{Experiments and results}
\label{experiments}
Experiments are based on the dataset that represents the blog/website posts of various world triathletes. Interestingly, almost all of the observed websites are organized as blogs. WordPress is the most popular platform, while a lot of blogs are hosted on BlogSpot or Wix.com platforms. Datasets were scraped and extracted automatically into a transaction database. In summary, the transaction database in the experiments consists of 4,271 feeds~\footnote{updated in December 2019}. 

In this database, the following elements of RSS feeds are included:
\begin{itemize}
\item title,
\item description,
\item link,
\item date, and
\item content.
\end{itemize}
It is worth  mentioning that some experiments were already conducted on the initial version of the database, while the initial findings were published in paper~\cite{fister2017deep}. In that paper, the active lifestyle of triathon athletes was analyzed, where deep analytic methods were proposed for analyzing these feeds. The results of the analysis were presented as social networks that serve as a basis for the decision-making process, from which the real triathlon trainer can extract some characteristics and information about the triathlon athlete`s lifestyle. 

However, the transaction database also has some limitations, referring especially to cleaning and removing  some inappropriate feeds. It seems that some websites were probably hacked, and some inappropriate bots posted, which caused the improper and strange content. Obviously, such feeds were also removed from the database. Similarly, the same procedure was applied also by preparing the database in the present study.

The parameter setting of the PSO for ARTM is proposed in Table~\ref{rules-stats},
\begin{table}
 \caption{Parameter settings of the PSO algorithm}
 \label{rules-stats}
 \begin{tabular}{|l|c|c|} 
 \hline
 Parameter & Abbreviation & Value\\ [0.5ex] 
 \hline\hline
 Population size & $\mathit{Np}$ & 200 \\ [0.5ex] 
 Social component & $C_1$ & 2.0 \\ [0.5ex] 
 Cognitive component & $C_2$ & 2.0 \\ [0.5ex] 
 Inertia weight & $w$ & 0.7 \\ [0.5ex] 
 Number of independent runs & $NUM\_RUNS$ & 5 \\ [0.5ex] 
 Number of fitness evaluations & $nFEs$ & 10,000 \\ [0.5ex] 
 \hline
\end{tabular}
\end{table}
from which it can be seen that the population size was limited to $\mathit{Np}=200$, the termination condition to the maximum number of fitness function evaluations $\mathit{nFEs}=10,000$, and there were five independent runs of the PSO algorithm conducted. 

\begin{figure}
  \centering
    \includegraphics[width=0.55\textwidth]{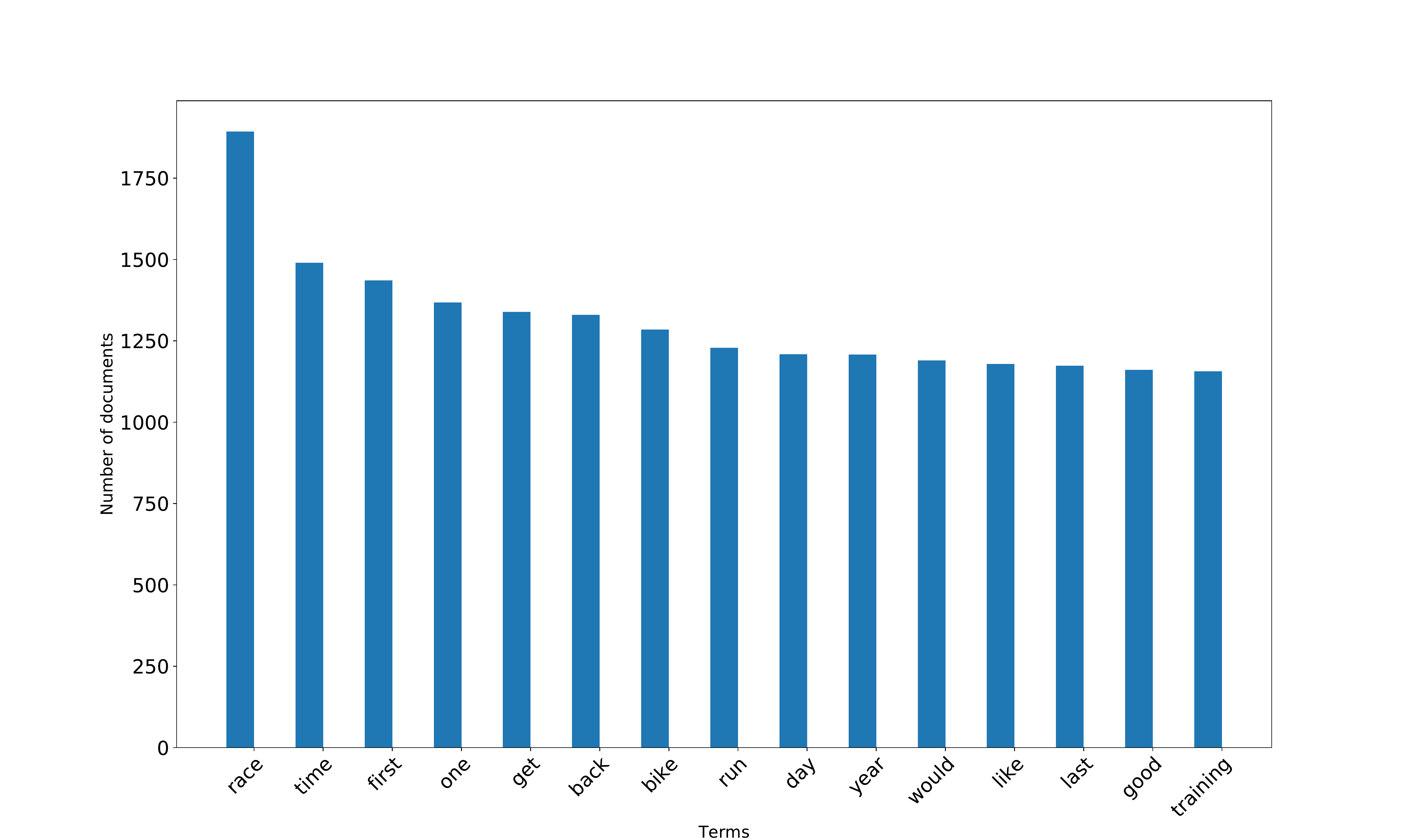}
     \caption{Histogram of the most frequent words in a database.}
     \label{fig:hist}
\end{figure}

Interestingly, the PSO algorithm is limited to process only the 1,000 most frequently occurring terms from the transaction database. The 15 of these are depicted in the histogram in Fig.~\ref{fig:hist}, 
where the terms are presented according to the number of occurrences. As can be seen from the histogram, the most frequently used term in the posted feeds is ''race''.

\subsection{The results}
The association rules are mined according to the number of terms $K$, that was varied in the interval $K\in[5,8]$ in steps of one. In this way, four instances were obtained, while the statistics of the mined rules are presented in Table~\ref{rules-stats},
\begin{table}
 \caption{Statistics of rules found on different $K$ settings}
 \label{rules-stats}
 \begin{tabular}{|c|c c c c||} 
 \hline
 $K$ & 5 & 6 & 7 & 8\\ [0.5ex] 
 \hline\hline
 No. Rules & 4594 & 1947 & 282 & 273\\ [0.5ex] 
 Avg Ant. & 1.693 & 2.776 & 2.148 & 2.520\\ [0.5ex] 
 Avg Cons. & 2.306 & 2.223 & 3.851 & 4.479\\ [0.5ex] 
 \hline
\end{tabular}
\end{table}
from which it is evident that the maximum number of rules is mined by the lower number of the parameter $K$. Interestingly, the average number of antecedents and consequences do not follow these trends, because the aforementioned values for $K=5$ are lower than the same for $K=8$.

Finally, the five  more interesting association rules mined using the PSO algorithm for ARTM are illustrated in Table~\ref{rez1}.
\begin{table}[htb]
\caption{Examples of some interesting solutions found by the proposed approach.}
\centering
\resizebox{\columnwidth}{!}{%
\begin{tabular}{|r|r|r|}
\hline
Rule & Antecedent & Consequence\\ \hline
1 & \makecell{amazing $\wedge$ ride $\wedge$ next} & \makecell{running $\wedge$ hurt $\wedge$ hopefully $\wedge$ fine} \\ \hline
2 & \makecell{championship $\wedge$ skills} & \makecell{race $\wedge$ technical} \\ \hline
3 & \makecell{great} & \makecell{year $\wedge$ news $\wedge$ mph $\wedge$ course $\wedge$ start $\wedge$ always} \\ \hline
4 & \makecell{one $\wedge$ race} &  \makecell{hard $\wedge$ bike $\wedge$ finish $\wedge$ week $\wedge$ amount} \\ \hline
5 & \makecell{triathlete $\wedge$ people} & \makecell{right $\wedge$ family $\wedge$ sprint} \\ \hline
\end{tabular}}
\label{rez1} 
\normalsize
\end{table} 
The meaning of these mined rules can be interpreted as follows: The first rule is referring to the relationship between cycling and running in a triathlon. From this rule, it is evident that, if athletes ride the cycles well, they are also good in running. The second rule asserts that the athlete who wants to be the champion, needs also to train some technical skills. The third rule is added intentionally by us  to show that the interpretation of some rules is not so easy. The fourth rule is referring to the hard cycling courses, while the last one exposes the importance of family support in triathlon racing.   

\section{Conclusion}
\label{conclusion}
This paper deals with the problem of Association Rule Text Mining, as well as proposing a new method for solving this problem using the stochastic population-based nature-inspired metaheuristics. The proposed PSO-ARTM method was evaluated on a transaction database consisting of the website/blog feeds of many world triathlon athletes. The results suggest that we can infer some useful information from these blogs.

Actually, we posted three questions at the beginning of the experimental study, and revealed that: (1) The stochastic population-based nature-inspired metaheuristics are suitable tools for solving  ARTM, (2) The interpretation of the discovered associated rules is not trivial, especially for rules with either one antecedent or one consequence, and (3) There is a bright way for the future development of these algorithms.

The last finding is justified as follows: Indeed, the future opens a broad path for a further enhancement of the method. For instance, the development of a better evaluation function or development of a multi-objective version of these methods might be one of the first steps for its improvement. Nevertheless, the evaluation of this method on the other well-known transaction databases could also represent a big challenge for the future.


\begin{thebibliography}{13}


\ifx \showCODEN    \undefined \def \showCODEN     #1{\unskip}     \fi
\ifx \showDOI      \undefined \def \showDOI       #1{#1}\fi
\ifx \showISBNx    \undefined \def \showISBNx     #1{\unskip}     \fi
\ifx \showISBNxiii \undefined \def \showISBNxiii  #1{\unskip}     \fi
\ifx \showISSN     \undefined \def \showISSN      #1{\unskip}     \fi
\ifx \showLCCN     \undefined \def \showLCCN      #1{\unskip}     \fi
\ifx \shownote     \undefined \def \shownote      #1{#1}          \fi
\ifx \showarticletitle \undefined \def \showarticletitle #1{#1}   \fi
\ifx \showURL      \undefined \def \showURL       {\relax}        \fi
\providecommand\bibfield[2]{#2}
\providecommand\bibinfo[2]{#2}
\providecommand\natexlab[1]{#1}
\providecommand\showeprint[2][]{arXiv:#2}

\bibitem[\protect\citeauthoryear{Alatas, Akin, and Karci}{Alatas
  et~al\mbox{.}}{2008}]%
        {alatas2008modenar}
\bibfield{author}{\bibinfo{person}{Bilal Alatas}, \bibinfo{person}{Erhan Akin},
  {and} \bibinfo{person}{Ali Karci}.} \bibinfo{year}{2008}\natexlab{}.
\newblock \showarticletitle{MODENAR: Multi-objective differential evolution
  algorithm for mining numeric association rules}.
\newblock \bibinfo{journal}{{\em Applied Soft Computing\/}}
  \bibinfo{volume}{8}, \bibinfo{number}{1} (\bibinfo{year}{2008}),
  \bibinfo{pages}{646--656}.
\newblock


\bibitem[\protect\citeauthoryear{Altay and Alatas}{Altay and Alatas}{2019}]%
        {altay2019performance}
\bibfield{author}{\bibinfo{person}{Elif~Varol Altay} {and}
  \bibinfo{person}{Bilal Alatas}.} \bibinfo{year}{2019}\natexlab{}.
\newblock \showarticletitle{Performance analysis of multi-objective artificial
  intelligence optimization algorithms in numerical association rule mining}.
\newblock \bibinfo{journal}{{\em Journal of Ambient Intelligence and Humanized
  Computing\/}} (\bibinfo{year}{2019}), \bibinfo{pages}{1--21}.
\newblock


\bibitem[\protect\citeauthoryear{Antonie and Zaiane}{Antonie and
  Zaiane}{2002}]%
        {antonie2002text}
\bibfield{author}{\bibinfo{person}{M-L Antonie} {and} \bibinfo{person}{Osmar~R
  Zaiane}.} \bibinfo{year}{2002}\natexlab{}.
\newblock \showarticletitle{Text document categorization by term association}.
  In \bibinfo{booktitle}{{\em 2002 IEEE International Conference on Data
  Mining, 2002. Proceedings.}} IEEE, \bibinfo{pages}{19--26}.
\newblock


\bibitem[\protect\citeauthoryear{Boytcheva}{Boytcheva}{2018}]%
        {boytcheva2018indirect}
\bibfield{author}{\bibinfo{person}{Svetla Boytcheva}.}
  \bibinfo{year}{2018}\natexlab{}.
\newblock \showarticletitle{Indirect Association Rules Mining in Clinical
  Texts}. In \bibinfo{booktitle}{{\em International Conference on Artificial
  Intelligence: Methodology, Systems, and Applications}}. Springer,
  \bibinfo{pages}{36--47}.
\newblock


\bibitem[\protect\citeauthoryear{Brownlee}{Brownlee}{2017}]%
        {brownlee2014machine}
\bibfield{author}{\bibinfo{person}{Jason Brownlee}.}
  \bibinfo{year}{2017}\natexlab{}.
\newblock \showarticletitle{Machine learning mastery}.
\newblock \bibinfo{journal}{{\em URL:
  http://machinelearningmastery.com/gentle-introduction-bag-words-model/\/}}
  (\bibinfo{year}{2017}).
\newblock


\bibitem[\protect\citeauthoryear{Fister~Jr., Fister, Rauter, Mlakar, Brest, and
  Fister}{Fister~Jr. et~al\mbox{.}}{2017}]%
        {fister2017deep}
\bibfield{author}{\bibinfo{person}{Iztok Fister~Jr.},
  \bibinfo{person}{Du{\v{s}}an Fister}, \bibinfo{person}{Samo Rauter},
  \bibinfo{person}{Uro{\v{s}} Mlakar}, \bibinfo{person}{Janez Brest}, {and}
  \bibinfo{person}{Iztok Fister}.} \bibinfo{year}{2017}\natexlab{}.
\newblock \showarticletitle{Deep analytics based on triathlon athletes' blogs
  and news}. In \bibinfo{booktitle}{{\em 23rd International Conference on Soft
  Computing}}. Springer, \bibinfo{pages}{279--289}.
\newblock


\bibitem[\protect\citeauthoryear{Fister~Jr., Iglesias, Galvez, Del~Ser, Osaba,
  and Fister}{Fister~Jr. et~al\mbox{.}}{2018}]%
        {Fister2018ARM}
\bibfield{author}{\bibinfo{person}{Iztok Fister~Jr.}, \bibinfo{person}{Andres
  Iglesias}, \bibinfo{person}{Akemi Galvez}, \bibinfo{person}{Javier Del~Ser},
  \bibinfo{person}{Eneko Osaba}, {and} \bibinfo{person}{Iztok Fister}.}
  \bibinfo{year}{2018}\natexlab{}.
\newblock \showarticletitle{Differential Evolution for Association Rule Mining
  Using Categorical and Numerical Attributes}. In \bibinfo{booktitle}{{\em
  International Conference on Intelligent Data Engineering and Automated
  Learning}}. \bibinfo{pages}{79--88}.
\newblock


\bibitem[\protect\citeauthoryear{Kamruzzaman, Haider, and Hasan}{Kamruzzaman
  et~al\mbox{.}}{2010}]%
        {kamruzzaman2010text}
\bibfield{author}{\bibinfo{person}{SM Kamruzzaman}, \bibinfo{person}{Farhana
  Haider}, {and} \bibinfo{person}{Ahmed~Ryadh Hasan}.}
  \bibinfo{year}{2010}\natexlab{}.
\newblock \showarticletitle{Text classification using association rule with a
  hybrid concept of naive Bayes classifier and genetic algorithm}.
\newblock \bibinfo{journal}{{\em arXiv preprint arXiv:1009.4976\/}}
  (\bibinfo{year}{2010}).
\newblock


\bibitem[\protect\citeauthoryear{Kennedy and Eberhart}{Kennedy and
  Eberhart}{1995}]%
        {kennedy1995particle}
\bibfield{author}{\bibinfo{person}{J Kennedy} {and} \bibinfo{person}{R
  Eberhart}.} \bibinfo{year}{1995}\natexlab{}.
\newblock \showarticletitle{{Particle swarm optimization}}. In
  \bibinfo{booktitle}{{\em Neural Networks, 1995. Proceedings., IEEE
  International Conference on}}, Vol.~\bibinfo{volume}{4}. IEEE,
  \bibinfo{pages}{1942--1948}.
\newblock


\bibitem[\protect\citeauthoryear{Kulkarni and Kulkarni}{Kulkarni and
  Kulkarni}{2016}]%
        {kulkarni2016knowledge}
\bibfield{author}{\bibinfo{person}{Manasi Kulkarni} {and}
  \bibinfo{person}{Sagar Kulkarni}.} \bibinfo{year}{2016}\natexlab{}.
\newblock \showarticletitle{Knowledge Discovery in Text Mining using
  Association Rule Extraction}.
\newblock \bibinfo{journal}{{\em International Journal of Computer
  Applications\/}} \bibinfo{volume}{143}, \bibinfo{number}{12}
  (\bibinfo{year}{2016}), \bibinfo{pages}{30--35}.
\newblock


\bibitem[\protect\citeauthoryear{Manimaran and Velmurugan}{Manimaran and
  Velmurugan}{2013}]%
        {manimaran2013survey}
\bibfield{author}{\bibinfo{person}{J Manimaran} {and} \bibinfo{person}{T
  Velmurugan}.} \bibinfo{year}{2013}\natexlab{}.
\newblock \showarticletitle{A survey of association rule mining in text
  applications}. In \bibinfo{booktitle}{{\em 2013 IEEE International Conference
  on Computational Intelligence and Computing Research}}. IEEE,
  \bibinfo{pages}{1--5}.
\newblock


\bibitem[\protect\citeauthoryear{Qodmanan, Nasiri, and Minaei-Bidgoli}{Qodmanan
  et~al\mbox{.}}{2011}]%
        {qodmanan2011multi}
\bibfield{author}{\bibinfo{person}{Hamid~Reza Qodmanan}, \bibinfo{person}{Mahdi
  Nasiri}, {and} \bibinfo{person}{Behrouz Minaei-Bidgoli}.}
  \bibinfo{year}{2011}\natexlab{}.
\newblock \showarticletitle{Multi objective association rule mining with
  genetic algorithm without specifying minimum support and minimum confidence}.
\newblock \bibinfo{journal}{{\em Expert Systems with applications\/}}
  \bibinfo{volume}{38}, \bibinfo{number}{1} (\bibinfo{year}{2011}),
  \bibinfo{pages}{288--298}.
\newblock


\bibitem[\protect\citeauthoryear{Rahman, Sohel, Naushad, and
  Kamruzzaman}{Rahman et~al\mbox{.}}{2010}]%
        {rahman2010text}
\bibfield{author}{\bibinfo{person}{Chowdhury~Mofizur Rahman},
  \bibinfo{person}{Ferdous~Ahmed Sohel}, \bibinfo{person}{Parvez Naushad},
  {and} \bibinfo{person}{SM Kamruzzaman}.} \bibinfo{year}{2010}\natexlab{}.
\newblock \showarticletitle{Text classification using the concept of
  association rule of data mining}.
\newblock \bibinfo{journal}{{\em arXiv preprint arXiv:1009.4582\/}}
  (\bibinfo{year}{2010}).
\newblock

\end{thebibliography}
\end{document}